# A HIERARCHICAL DIRICHLET PROCESS MIXTURE MODEL FOR HAPLOTYPE RECONSTRUCTION FROM MULTI-POPULATION DATA[1]

By Kyung-Ah Sohn and Eric P. Xing [2]

*Carnegie Mellon University*

The perennial problem of "how many clusters?" remains an issue of substantial interest in data mining and machine learning communities, and becomes particularly salient in large data sets such as populational genomic data where the number of clusters needs to be relatively large and open-ended. This problem gets further complicated in a *co-clustering* scenario in which one needs to solve multiple clustering problems simultaneously because of the presence of common centroids (e.g., ancestors) shared by clusters (e.g., possible descents from a certain ancestor) from different multiple-cluster samples (e.g., different human subpopulations). In this paper we present a hierarchical nonparametric Bayesian model to address this problem in the context of multi-population haplotype inference.

Uncovering the haplotypes of single nucleotide polymorphisms is essential for many biological and medical applications. While it is uncommon for the genotype data to be pooled from multiple ethnically distinct populations, few existing programs have explicitly leveraged the individual ethnic information for haplotype inference. In this paper we present a new haplotype inference program, *Haploi*, which makes use of such information and is readily applicable to genotype sequences with thousands of SNPs from heterogeneous populations, with competent and sometimes superior speed and accuracy comparing to the state-of-the-art programs. Underlying *Haploi* is a new haplotype distribution model based on a nonparametric Bayesian formalism known as the *hierarchical Dirichlet process*, which represents a tractable surrogate to the coalescent process. The proposed model

Received August 2008; revised November 2008.

[1]Supported by the National Science Foundation under Grant No. CCF-0523757 and by the Pennsylvania Department of Health's Health Research Program under Grant No. 2001NF-Cancer Health Research Grant ME-01-739.

[2]Supported by an NSF CAREER Award, under Grant No. DBI-054659 and an Alfred P. Sloan Research Fellowship in Computer Science.

is exchangeable, unbounded, and capable of coupling demographic information of different populations. It offers a well-founded statistical framework for posterior inference of individual haplotypes, the size and configuration of haplotype ancestor pools, and other parameters of interest given genotype data.

**1. Introduction.** Recent experimental advances have led to an explosion of data that document genetic variations within and between populations. For example, the International SNP Map Working Group (2001) has reported the identification and mapping of 1.4 million *single nucleotide polymorphisms* (SNPs) from the genomes of four different human populations. These data pose challenging inference problems whose solutions could shed light on the evolutionary history of human population and the genetic basis of disease propensities [Chakravarti (2001), Clark (2003)].

SNPs represent the largest class of individual differences in DNA. A SNP refers to the existence of two possible nucleotide bases from $\{A, C, G, T\}$ at a chromosomal locus in a population; each variant, denoted as 1 or 0, is called an *allele*. A *haplotype* refers to the joint allelic identities of a contiguous list of polymorphic loci within a study region on a given chromosome. Diploid organisms such as human beings have two haplotypes in each individual, one maternal copy and one paternal copy. When the parental chromosomes come in pairs, two haplotypes go together and make up a *genotype* which consists of the list of allele-pairs at every locus. More precisely, a genotype is resulted from a pair of haplotypes by omitting the information regarding the specific association of each allele with one of the two chromosomes—its *phase*, at every locus. The problem of haplotype inference, which is the focus of this paper, concerns determining which phase reconstruction among many alternatives is more plausible. Common biological methods for assaying genotypes typically do not provide phase information for individuals with heterozygous genotypes at multiple loci. Although phase can be obtained at a considerably higher cost via molecular haplotyping [Patil et al. (2001)], or sometimes from analysis of trios [Hodge, Boehnke and Spence (1999)], it is desirable to develop automatic and robust *in silico* methods for reconstructing haplotypes from the inexpensive genotype data.

Key to the inference of individual haplotypes based on a given genotype sample is the formulation and tractability of the marginal distribution of the haplotypes of the study population. Consider the set of haplotypes, denoted as $H = \{h_1, h_2, \ldots, h_{2n}\}$, of a random sample of $2n$ chromosomes of $n$ individuals. Under common genetic arguments, the ancestral relationships among the sample back to its most recent common ancestor (MRCA) can be described by a genealogical tree known as the *coalescent*; computing the $P(H)$ involves a marginalization over all possible coalescent trees compatible with the sample, which is widely known to be intractable. Li and Stephens (2003) suggested to approximate $P(H)$ by a *Product of Approximate Conditionals* (PAC). The PAC model tries to incorporate a desirable evolution assumption known as the *parental-dependent-mutation* (PDM) by modeling each $h_i$ as the progeny of a randomly-chosen existing haplotype, and it forms the basis of the PHASE program, which has set the state-of-the-art benchmark in haplotype inference. However, the PAC model implicitly assumes existence of an ordering in the haplotype sample, therefore, the resulting likelihood is not *exchangeable* as one would expect for the true $P(H)$. Moreover, since PAC involves no explicit ancestral genealogy over existing haplotypes, certain latent demographic information such as founding haplotypes and their mutation rates are not directly captured in the model.

The finite mixture models represent another class of haplotype models that rely very little on demographic and genetic assumptions of the sample [Excoffier and Slatkin (1995), Niu et al. (2002), Kimmel and Shamir (2004), Zhang, Niu and Liu (2006)]. Under such a model, haplotypes are treated as latent variables associated with specific frequencies, and the haplotype inference problem can be viewed as a *missing value inference* and *parameter estimation* problem, for which numerous statistical inference approaches have been developed, such as the maximum likelihood approaches via the EM algorithm [Excoffier and Slatkin (1995), Hawley and Kidd (1995), Long, Williams and Urbanek (1995), Fallin and Schork (2000)], and a number of parametric Bayesian inference methods based on Markov chain Monte Carlo (MCMC) sampling [Niu et al. (2002), Zhang, Niu and Liu (2006)]. However, this class of methods has rather severe computational requirements in that a probability distribution must be maintained on a (large) set of *possible* haplotypes. Indeed, the size of the haplotype pool, $K$, which reflects the diversity of the genome, is unknown for any given population data and needs to be inferred. There is a plethora of combinatorial algorithms based on various hypotheses, such as the "parsimony" principles that offer control over the complexity of the inference problem [see Gusfield (2004) for an excellent survey]. On the other hand, most current methods based on statistical inference employ computationally expensive techniques such as cross validation or model-selection to address the issue of ancestral-space uncertainty [Scheet and Stephens (2006)].

Indeed, the uncertainty regarding the size of the haplotype pool is an instance of the perennial problem of "how many clusters?" in the clustering literature. The problem is particularly salient in large data sets where the number of clusters needs to be relatively large and open-ended—exactly the scenario in population genomic analysis. In Xing, Sharan and Jordan (2004) we have proposed a nonparametric Bayesian model, specifically the *Dirichlet process* (DP) [Blackwell and MacQueen (1973), Ferguson (1973)], which provides a prior and posterior distribution for mixture models with unbounded numbers of mixture components. Recently, substantial efforts have also been made to speed up haplotype inference on large scale data. Notable programs include Beagle [Browning and Browning (2007)], which uses a localized haplotype model based on variable-length Markov chains, and MACH [Li and Abecasis (2006)], a fast version of PHASE.

These progresses notwithstanding, it is noteworthy that most progresses being made so far on approximating the $P(H)$ and $K$, and on dealing with long SNP sequences, do not explicitly leverage the potentially useful side information such as the genetic origins of individuals in a population sample. In particular, statistical models developed so far are inadequate for addressing the multi-population haplotype sharing problems concerned in this paper. Consider, for example, a genetic demography study, in which one seeks to uncover ethnic- or geographic-specific genetic patterns based on a sparse census of multiple populations. In particular, suppose that we are given a sample that can be divided into a set of subpopulations, for example, African, Asian and European. When conducting haplotype inference on such data, we may not only want to discover the sets of haplotypes within each subpopulation, but also which haplotypes are shared between subpopulations, and what their frequencies are. Empirical and theoretical evidence suggests that an early split of an ancestral population following a populational bottleneck (e.g., due to sudden migration or environmental changes) can lead to subpopulation-specific genetic diversity, which causes ancient haplotypes (that have higher variability) to be shared among different subpopulations, and unique modern haplotypes (that are more strictly conserved) to be instantiated and inherited in different subpopulations [Pritchard (2001)]. This structure is analogous to a co-clustering scenario in which different groups comprising multiple clusters may share clusters with common centroids (e.g., different news topics may share some common key words). The implication of this phenomenon on haplotype reconstruction has not been thoroughly investigated.

A naive solution to the aforementioned problem would be to infer haplotypes separately in the subpopulations. This is clearly suboptimal, however, because it may unnecessarily fragment the data, and may lead to unrobust estimation of demographic parameters. In particular, for rare haplotypes that are present in a small number of individuals (e.g., one or two) in each population but overall still have many bearers across all populations, the estimation of their founders (i.e., the centroid) should take into account of these bearers in all populations jointly, rather than being based on each population separately. Essentially, what we want is a model to solving multiple clustering problems simultaneously. In this paper we describe a new haplotype model based on a *hierarchical Dirichlet process* (HDP) [Teh et al. (2006), Xing et al. (2006)], which directly address this issue. An HDP over a measurable space $(\Phi, \mathcal{B})$ specifies a set of coupled random distributions

$\{\mathcal{Q}_1, \mathcal{Q}_2, \ldots, \mathcal{Q}_J\}$ on $\Phi \equiv \mathcal{E} \times \mathcal{A}$, where $\mathcal{E} = [0, 1]$ and $\mathcal{A} = \{0, 1\}^T$ denote the space of the mutation rates and joint allele configurations, respectively, of the *ancestral haplotypes* of $T$ SNP loci. Each $\mathcal{Q}_j$ is a population-specific Dirichlet process (DP) [Blackwell and MacQueen (1973), Ferguson (1973)] which defines a nonparametric prior over the ancestral haplotypes and their frequencies of being inherited within the population, and thereby induces a Dirichlet process mixture (DPM) model [Antoniak (1974)] for all the individual haplotypes in that population. As detailed in Section 2, to allow every ancestral haplotype in a particular population to also have nonzero probability of being inherited in a different population (albeit with different frequencies), a hyper-prior $\mathcal{Q}_0$, which is also a Dirichlet process and therefore discrete on $\Phi$, is used to define the base measures of each population-specific $\mathcal{Q}_j$, ensuring that they are all realized on a common set of supports (i.e., ancestors) in $\Phi$. Our model differs from other methods reviewed earlier in the following ways: (1) Instead of resorting to empirical assumptions or model selection over the number of population haplotypes, we introduce a nonparametric prior over haplotype ancestors, which facilitates posterior inference of the haplotypes in an "open" state space accommodating arbitrary sample size. (2) Our model explicitly exploits the subpopulation labels and potentially latent genetic demographic structures to improve haplotyping accuracy. (3) Our model captures similar genetic properties as those emphasized in Stephens, Smith and Donnelly (2001), including the parent-dependent-mutations, but with an *exchangeable* likelihood function.

We have developed an efficient MCMC-based software program *Haploi*, based on our proposed model, and using a variant of the Partition–Ligation scheme by Niu et al. (2002) to handle complexity explosion due to long input sequence. It can be readily applicable to multi-population genotype sequences, at a time–cost often at least two-orders of magnitude less than that of the state-of-the-art PHASE program, with competitive performance. We also show that *Haploi* can significantly outperform other popular haplotype inference algorithms on both simulated and real short SNPs data.

**2. The statistical model.** Our narration below starts with a basic Dirichlet process mixture model for a simple demographic scenario, where we ignore individual subpopulation labels and assume absence of recombination in the sample. Then we describe the hierarchical Dirichlet process mixture for haplotypes from multiple populations in detail. There is an interesting connection of the DPM-based methods to the Wright–Fisher model and Kingman's coalescent with an infinitely-many-alleles (IMA) mutation process for allele evolution, which we will briefly discuss.

2.1. *Dirichlet process mixture for haplotypes.* A random probability measure $\mathcal{Q}$ on a measurable space $(\Phi, \mathcal{B})$ is generated by a Dirichlet process

DP$(\tau, Q_0)$ if for every measurable partition $B_1, \ldots, B_k$ of the sample space $\Phi$, the vector of random probabilities $\mathcal{Q}(B_i)$ follows a finite dimensional Dirichlet distribution: $(\mathcal{Q}(B_1), \ldots, \mathcal{Q}(B_k)) \sim \text{Dir}(\tau Q_0(B_1), \ldots, \tau Q_0(B_k))$, where $\tau > 0$ denotes a *scaling parameter* and $Q_0$ denotes a *base measure* defined on $(\Phi, \mathcal{B})$ [Ferguson (1973)].

Samples from a DP tend to cluster around the distinct-valued atoms, which offers a salient way for us to group haplotypes around common ancestors. This property is best reflected in the constructive definition of the DP based on the following Pólya urn scheme [Blackwell and MacQueen (1973)]. Having observed $n$ samples with values $(\phi_1, \ldots, \phi_n)$ from DP$(\tau, Q_0)$, the conditional distribution of the value of the $(n+1)$th sample is given by

$$(1) \qquad \phi_{n+1}|\phi_1, \ldots, \phi_n, \tau, Q_0 \sim \sum_{k=1}^{K} \frac{n_k}{n+\tau} \delta_{\phi_k^*}(\cdot) + \frac{\tau}{n+\tau} Q_0(\cdot),$$

where $n_k$ denotes the number of samples with value $\phi_k^*$, and $K$ denotes the number of unique values in the $n$ samples drawn so far. This expression means that each new sample has positive probability of being equal to an existing unique value in the drawn samples, and, moreover, the probability is proportional to the *occupancy number* $n_k$ of the unique values, creating a clustering effect. The cluster cardinality $K$ is a random integer that is only bounded by the sample size, rather than being pre-specified.

To model a haplotype population $H$ that is genetically homogeneous, one can assume that $H$ is originated from a size-unknown group of distinct ancestral haplotypes (i.e., founders), and associate each unique value $\phi_k^*$ from a DP with a possible founder and its mutation probability, that is, $\{a_k, \theta_k\}$. Relating every drawn sample $\phi_i$ to a modern individual haplotype via a conditional likelihood function, we arrive at a DP mixture model [Antoniak (1974), Escobar and West (1995)] for $P(H)$ as described in Xing, Sharan and Jordan (2004), which is briefly recapitulated in the sequel for self-containedness and to introduce necessary notation for subsequent exposition of the hierarchical DP mixture for multi-population data. Specifically, write $H_{i_e} = [H_{i_e,1}, \ldots, H_{i_e,T}]$, where the sub-subscript $e \in \{0,1\}$ denotes the two possible parental origins (i.e., paternal and maternal), for a haplotype over $T$ contiguous SNPs from individual $i$, and let $G_i = [G_{i,1}, \ldots, G_{i,T}]$ denote the genotype of these SNPs of individual $i$. Let $A_k = [A_{k,1}, \ldots, A_{k,T}]$ denote an ancestor haplotype and $\theta_k$ denote the *mutation rate* of ancestor $k$, and let $C_i$ denote an *inheritance variable* that specifies the ancestor of haplotype $H_i$. $P_h(H|A, \theta)$ represents the *inheritance model* according to which individual haplotypes are derived from a founder, and $P_g(G|H_0, H_1)$ indicates the *genotyping model* via which noisy observations of the geno-types are related to the haplotypes. A DPM defines the following generative scheme:

- Draw first haplotype:

$$a_1, \theta_1 | \mathrm{DP}(\tau, Q_0) \sim Q_0(\cdot),$$

  sample the 1st founder (and its mutation rate);

$$h_1 \sim P_h(\cdot | a_1, \theta_1),$$

  sample the 1st haplotype from an inheritance model defined on the 1st founder;
- for subsequent haplotypes:
  – sample the founder indicator for the $i$th haplotype:

$$c_i | \mathrm{DP}(\tau, Q_0) \sim \begin{cases} P(c_i = c_j \text{ for some } j < i | c_1, \ldots, c_{i-1}) = \dfrac{n_{c_j}}{i - 1 + \alpha}, \\ P(c_i \neq c_j \text{ for all } j < i | c_1, \ldots, c_{i-1}) = \dfrac{\alpha}{i - 1 + \alpha}, \end{cases}$$

  where $n_{c_i}$ is the *occupancy number* of founder $a_{c_i}$.
  – sample the founder of haplotype $i$:

$$a_{c_i}, \theta_{c_i} | \mathrm{DP}(\tau, Q_0) \begin{cases} = \{a_{c_j}, \theta_{c_j}\}, & \text{if } c_i = c_j \text{ for some } j < i, \\ \sim Q_0(a, \theta), & \text{if } c_i \neq c_j \text{ for all } j < i, \end{cases}$$

  – sample the haplotype according to its founder:

$$h_i | c_i \sim P_h(\cdot | a_{c_i}, \theta_{c_i}).$$

- sample all genotypes (according to a mapping between haplotype index $i$ and allele index $i_e$):

$$g_i | h_{i_0}, h_{i_1} \sim P_g(\cdot | h_{i_0}, h_{i_1}).$$

Here, $\{a_k, \theta_k\}$ corresponds to the set of *mixture components*, and the DP is used as the prior over the components in an unbounded ancestral space. This prior requires no specification of the size of the ancestor pool.

2.2. *Hierarchical DP mixture for multi-population haplotypes.* Now consider the case where there exist multiple ethnic or geographic populations. Instead of modeling these subpopulations independently by unrelated DPMs, we place all the population-specific DPMs under a common prior, such that the ancestors in any of the population-specific mixtures can be shared across all the mixtures, but the *weight* of an ancestral haplotype in each mixture is unique.

To tie population-specific DP mixtures together in this way, we employ a hierarchical DP (HDP) mixture model [Teh et al. (2006)], in which the base measures of the all population-specific DPMs admit a common discrete prior defined by another Dirichlet process $DP(\gamma, F)$. An HDP defines a distribution over a set of dependent random probability measures, $\{Q_j, j = 1, \ldots, J\}$, and another master random probability measure $Q_0$ that controls all the $Q_j$'s. Each $Q_j$ is a population specific DP with common (or population-specific) scaling parameter $\tau$, and a shared base measure defined by $Q_0$. Moreover, $Q_0$ itself follows a Dirichlet process $DP(\gamma, F)$. Following a hierarchical Pólya urn scheme, for $m_j$ random draws $\phi_j = \phi_{j,1}, \ldots, \phi_{j,m_j}$ from $Q_j$, we can derive the following conditional probability for $(\phi_{m_j} | \boldsymbol{\phi}_{-m_j})$ [Xing et al. (2006)], where the subscript $-m_j$ denotes the index set of all but the $m_j$th sample:

$$
\begin{aligned}
\phi_{m_j} | \boldsymbol{\phi}_{-m_j} &\sim \sum_{k=1}^{K} \frac{m_{j,k} + \tau n_k/(n-1+\gamma)}{m_j - 1 + \tau} \delta_{\phi_k^*}(\phi_{m_j}) \\
&\quad + \frac{\tau}{m_j - 1 + \tau} \frac{\gamma}{n - 1 + \gamma} F(\phi_{m_j}) \\
&= \sum_{k=1}^{K} \pi'_{j,k} \delta_{\phi_k^*}(\phi_{m_j}) + \pi'_{j,K+1} F(\phi_{m_j}),
\end{aligned}
\tag{2}
$$

where $n_k$ denotes the number of samples under $Q_0$ drawn from the global measure $F$ and equal to $\phi_k^*$, $m_{j,k}$ denotes the number of samples in the $j$th group which are equal to $\phi_k^*$, and $\pi'_{j,k} := \frac{m_{j,k} + \tau n_k/(n-1+\gamma)}{m_j - 1 + \tau}$, $\pi'_{j,K+1} = \frac{\tau}{m_j - 1 + \tau} \frac{\gamma}{n - 1 + \gamma}$. The vector $\vec{\pi}'_j = (\pi'_{j,1}, \pi'_{j,2}, \ldots)$ gives the a priori conditional probability of a new sample in group $j$. As shown later, this formula will be useful for implementing a Gibbs sampler for posterior inference under HDP mixtures.

Based on the HDP described above, we now define an HDP mixture (HDPM) model for the genotypes in $J$ populations. Elaborating on the notational scheme used earlier, let $G_i^{(j)} = [G_{i,1}^{(j)}, \ldots, G_{i,T}^{(j)}]$ denote the *genotype* of $T$ contiguous SNPs of individual $i$ from subpopulation group $j$, and let $H_{i_e}^{(j)} = [H_{i_e,1}^{(j)}, \ldots, H_{i_e,T}^{(j)}]$ denote a haplotype of individual $i$ from ethnic group $j$. The basic generative structure of multi-population genotypes under an HDPM is as follows, which is also illustrated graphically in Figure 1:

$$\mathcal{Q}_0(\phi_1, \phi_2, \ldots)|\gamma, F \sim \mathrm{DP}(\gamma, F),$$

sample a DP of founders for all populations;

$$\mathcal{Q}_j(\phi_1^{(j)}, \phi_2^{(j)}, \ldots)|\tau, Q_0 \sim \mathrm{DP}(\tau, \mathcal{Q}_0),$$

sample the DP of founders for each population;

$$\phi_{i_e}^{(j)}|\mathcal{Q}_j \sim \mathcal{Q}_j,$$

sample the founder of haplotype $i_e$ in population $j$;

$$h_{i_e}^{(j)}|\phi_{i_e}^{(j)} \sim P_h(\cdot|\phi_{i_e}^{(j)}),$$

sample haplotype $i_e$ in population $j$;

$$g_i^{(j)}|h_{i_0}^{(j)}, h_{i_1}^{(j)} \sim P_g(\cdot|h_{i_0}^{(j)}, h_{i_1}^{(j)}),$$

sample genotype $i$ in population $j$,

where the first three steps of sampling founder haplotypes follow the HDP scheme, the fourth step describes the mixture formulism, and the last step corresponds to the noisy genotyping model. Recall that in an HDP the base measure $\mathcal{Q}_0$ is a *random distribution* of the pool of haplotype founders and their associated mutation rates. It ensures that all the population-specific child DPs can be defined on a *common* unbounded pool of candidate founder patterns. The child DPs place different mass distributions, that is, a priori frequencies of haplotype founders, on this common support, in a population-specific fashion.

The base measure $F$ in the above generative process is defined as a distribution from which haplotype founders $\phi_k \equiv \{A_k, \theta_k\}$ are drawn. Thus, it is a joint measure on both $A$ and $\theta$. We let $F(A, \theta) = p(A)p(\theta)$, where $p(A)$ is uniform over all possible haplotypes and $p(\theta)$ is a beta distribution introducing a prior belief of low mutation rate. For generality, we assume $A_{k,t}$ and $H_{i,t}$ of every single locus take values from an allele set $\mathcal{A}$. For other building blocks of the HDPM model, we propose the following specifications.

2.2.1. *Haplotype inheritance model.* Omitting all but the locus index $t$, we can define our inheritance model to be a *single-locus mutation model* as follows [Xing, Sharan and Jordan (2004)]:

$$(3) \qquad P_h(h_t|a_t, \theta) = (1-\theta)^{\mathbb{I}(h_t = a_t)} \left(\frac{\theta}{|\mathcal{A}|-1}\right)^{\mathbb{I}(h_t \neq a_t)},$$

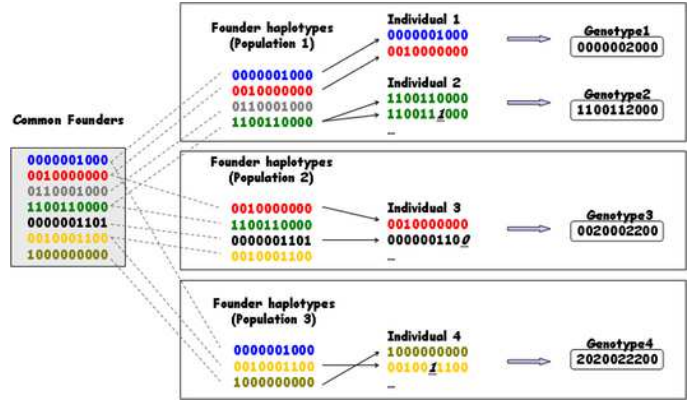

FIG. 1. *The haplotype–genotype generative process under HDPM, illustrated by an example concerning three populations. At the first level, all haplotype founders from different populations are drawn from a common pool via a Pólya urn scheme, which leads to the following effects: 1. the same founder can be drawn by either multiple populations (e.g., the red founder in population 1 and 2, and the blue one in population 1 and 3), or only a single population (e.g., the grey founder in population 1); 2. shared founders can have different frequencies of being inherited. Then at the second level, individual haplotypes were drawn from a population-specific founder pool also via a Pólya urn scheme, but this time through an inheritance model $P_h(\cdot|a_k)$ that allows mutations with respect to the founders, as indicated by the underscores at the mutated loci in the individual haplotypes. Finally, genotypes are related to the haplotype pairs of every individual via a noisy channel $P_g(\cdot|\cdot)$.*

where $\mathbb{I}(\cdot)$ is the indicator function. This model corresponds to a star genealogy resulting from infrequent mutations over a shared ancestor, and is widely used as an approximation to a full coalescent genealogy starting from the shared ancestor [e.g., Liu et al. (2001)].

Given this inheritance model, and under a beta prior $\text{Beta}(\alpha_h, \beta_h)$ for the mutation rate $\theta$, it can be shown that the marginal conditional distribution of a haplotype sample $\mathbf{h} = \{h_{i_e} : e \in \{0,1\}, i \in \{1, 2, \ldots, I\}\}$ takes the following form resulted from an integration of $\theta$ in the joint conditional:

$$(4) \quad p(\mathbf{h}|\mathbf{a}, \mathbf{c}) = \prod_{k=1}^{K} R(\alpha_h, \beta_h) \frac{\Gamma(\alpha_h + l_k)\Gamma(\beta_h + l'_k)}{\Gamma(\alpha_h + \beta_h + l_k + l'_k)} \left(\frac{1}{|\mathcal{A}| - 1}\right)^{l'_k},$$

where $R(\alpha_h, \beta_h) = \frac{\Gamma(\alpha_h + \beta_h)}{\Gamma(\alpha_h)\Gamma(\beta_h)}$, $l_k = \sum_{i,e,t} \mathbb{I}(h_{i_e,t} = a_{k,t})\mathbb{I}(c_{i_e} = k)$ is the number of alleles which are identical to the ancestral alleles, and $l'_k = \sum_{i,e,t} \mathbb{I}(h_{i_e} \neq a_{k,t})\mathbb{I}(c_{i_e} = k)$ is the total number of mutated alleles.

2.2.2. *Genotype observation model.* We assume that the genotype at a locus is determined by the paternal and maternal alleles of this site via the following genotyping model [Xing, Sharan and Jordan (2004)]:

$$(5) \quad P_g(g|h_{i_0,t}, h_{i_1,t}; \xi) = \xi^{\mathbb{I}(h=g)}[\mu_1(1-\xi)]^{\mathbb{I}(h\neq^1 g)}[\mu_2(1-\xi)]^{\mathbb{I}(h\neq^2 g)},$$

where $h \triangleq h_{i_0,t} \oplus h_{i_0,t}$ denotes the unordered pair of two actual SNP allele instances at locus $t$; "$\neq^1$" denotes set difference by exactly one element; "$\neq^2$" denotes set difference of both elements, and $\mu_1$ and $\mu_2$ are appropriately defined normalizing constants. Again we place a beta prior $\text{Beta}(\alpha_g, \beta_g)$ on $\xi$ for smoothing. Under the above model specifications, it is standard to derive the posterior distribution of each haplotype $H_{i_e}$ given all other haplotypes and all genotypes, and the posterior of any missing genotypes, by integrating out parameters $\theta$ or $\xi$ and resorting to the Bayes theorem, which enables a collapsed Gibbs sampling step where necessary.

2.2.3. *Hyperprior for scaling parameters.* To capture uncertainty over the scaling parameters, for example, $\gamma$, we use a vague inverse Gamma prior:

$$(6) \qquad p(\gamma^{-1}) \sim \mathcal{G}(1,1) \quad \Rightarrow \quad p(\gamma) \propto \gamma^{-2} \exp(-1/\gamma).$$

In general, the probability density function of an inverse Gamma distribution with shape parameter $\iota$ and scale parameter $\kappa$ is given as follows:

$$p(x; \iota, \kappa) = \frac{\kappa^\iota}{\Gamma(\iota)} x^{-\iota-1} \exp\left(\frac{-\kappa}{x}\right).$$

Under this prior, the posterior distribution of $\gamma$ depends only on the number of instances $n$, and the number of components $K$, but not on how the samples are distributed among the components:

$$(7) \qquad p(\gamma|k,n) \propto \frac{\gamma^{k-2} \exp(1/\gamma) \Gamma(\gamma)}{\Gamma(n+\gamma)}.$$

The distribution $p(\log(\gamma)|k,n)$ is log-concave, so we may efficiently generate independent samples from this distribution using adaptive rejection sampling [Rasmussen (2000)]. It is noteworthy that in an HDPM we need to define vague inverse Gamma priors also for the scaling parameters $\tau$ of population-specific DPs at the bottom level. We use a single concentration parameter $\tau$ for these DPs; it is also possible to allow separate concentration parameters for each of the lower-level DPs, possibly tied distributionally via a common hyperparameter.

2.3. *Posterior inference via Gibbs sampling.* Based on the two-level Pólya urn implementation of HDPM, an efficient MCMC algorithm, which is similar to the MCMC algorithms developed for DPM, can be derived to sample from the posterior associated with HDPM. Specifically, under a collapsed Gibbs sampling scheme where $\theta$ and $\xi$ are integrated out, the variables of interest are $C_{i_e,t}^{(j)}$, $A_{k,t}$, $H_{i_e,t}^{(j)}$, $\gamma$, and $\tau$, $\forall i,j,k,t,e$. The sampler alternates between three coupled stages. First, it samples the scaling parameters $\gamma$ and $\tau$ of the DPs, following the predictive distribution given by equation (7).

Then, it samples the $c_{i_e}^{(j)}$'s and $a_{k,t}$'s given the current values of the hidden haplotypes and the scaling parameters according to equations (8) and (9) (Appendix), respectively. Finally, given the current state of the ancestral pool, the ancestor assignment for each individual and the observed genotypes, it samples the $h_{i_e,t}^{(j)}$ variables according to equation (10). Details of the forms and derivations of the predictive distributions used for steps 2 and 3 are given in the Appendix.

2.4. *Population genetic implication of DP-based haplotype models.* There is an interesting connection between the Dirichlet process models and the well-known coalescent process theory underlying population genetic evolution [Kingman (1982)]. It can be shown that an *infinitely-many-alleles* (IMA) model with rate $\tau/2$ on an $n$-coalescent extends haplotype lineages on the coalescent tree according to the following law: with probability $\tau/(n-1+\tau)$, it instantiates a new haplotype, and with probability $(n-1)/(n-1+\tau)$, it replicates an existing haplotype lineage [Hoppe (1984)]. This is identical to the Pólya urn scheme described in equation (1) with scaling parameters $\tau$ and uniform base distribution over $\mathcal{A}$, a Dirichlet process $DP(\tau, \text{Uniform})$.

There is a mapping between the distinct founders $\phi_k^* \equiv \{a_k, \theta_k\}, \forall k$ arising from a DP, to the novel haplotypes generated according to IMA on a coalescent tree at the birth of every new lineage. However, since these founders are independently draw, from the base measure, a basic DP cannot capture relationships between different founding haplotypes in a population.

The *parental-dependent-mutation* model posits that, in a sequential generation process of haplotypes, if the next haplotype does not match exactly with an existing haplotype, it will tend to differ by a small number of mutations from an existing one, rather than be completely different. Under a DP mixture, modern individual haplotypes $h_i$ are marginally dependent, because similar but nonidentical haplotypes can be grouped around possible founders according to an inheritance model $P_h(H|A,\theta)$ that permits further changes on top on founders. As discussed later, this leads to an exchangeable $P(H)$ that captures the effect of parent-dependent mutations.

**3. Partition-ligation and the Haploi program.** As for most haplotype inference models proposed in the literature, the state space of the proposed HDPM model scales exponentially with the length of the genotype sequence and, therefore, it cannot be directly applied to genotype data containing hundreds or thousands of SNPs. To deal with haplotypes with a large number of linked SNPs, Niu et al. (2002) proposed a divide-and-conquer heuristic known as Partition–Ligation (PL), which was adopted by a number of haplotype inference algorithms including PL-EM [Qin, Niu, and Liu (2002)], PHASE [Stephens, Smith and Donnelly (2001), Li and Stephens (2003)] and CHB [Zhang, Niu and Liu (2006)]. We equipped the HDPM model with a variant of the PL heuristic, and present a new tool, *Haploi* for *haplotype inference* of multiple population genotype data over long SNPs sequences.

The original PL-scheme in Niu et al. (2002) first divides the entire sequence into disjoint short blocks and reconstructs haplotypes within each block. Then pairs of blocks are recursively ligated into larger (nonoverlapping) haplotypes via Gibbs sampling under a fixed-dimensional Dirichlet prior over the frequencies of the ligated haplotype in the *product space* (or a subset) of all the "atomistic haplotypes" of every pair of blocks. This bottom-up approach can recover haplotypes of every individual either hierarchically or progressively. However, this PL scheme does not scale well to long sequences because the number of possible haplotypes in the product space can quickly become intractable as the size of the nonoverlapping blocks to be ligated grows multiplicatively during the iteration. Unlike their approach, our PL-scheme generates partially overlapping intermediate blocks from smaller blocks phased at the lower level. The pairs of overlapping blocks are recursively merged into larger ones by leveraging the redundancy of information from overlapping regions, as well as overall parsimonious criteria. Empirically we found that this strategy can lead to a significant reduction of the size of the haplotype search space for long genotypes, and therefore facilitates a more efficient inference algorithm.

Figure 2 outlines the PL-procedure adopted by *Haploi*, which can be divided into three steps. In step 1 we begin by partitioning given genotype sequences into $L$ short blocks of length $T$ [e.g., $T \leq 10$ as suggested in Niu et al. (2002)]. Then we phase each atomistic block using the proposed HDPM (Figure 2, step 1). By doing this, we obtain all the individual haplotypes and also the population haplotype pool (i.e., founders) for each block. In the next step we ligate every pair of neighboring blocks. Naively the candidate population haplotype pool for the ligated segment can be a Cartesian product of the haplotype pools in neighboring blocks. But such an unconstrained product is in fact unnecessary. Since each individual harbors only two possible haplotypes within each block, for each pair of adjacent blocks, we can impute at most four new stitched haplotypes from an individual, but, in practice, we get much fewer because an individual can be homozygous on one or both blocks and the stitched haplotypes may have been imputed already from earlier individuals; also, not all combinations of haplotypes in the two pools are necessary because some combinations may never exist in any individual. We pool such stitched haplotypes imputed from all individuals, which usually leads to only a small subset of the Cartesian product of the two haplotype pools. Then based on a finite dimensional Dirichlet prior over the candidate pool, we do Gibbs sampling as in Niu et al.'s PL scheme to obtain individual haplotypes for each overlapping $2T$ region. Essentially, our procedure produces a more parsimonious set of population haplotypes by using an individual-based population haplotype imputation scheme. In addition, comparing to the ligation in Niu et al.'s scheme, we stitch every neighboring pair of blocks [$i$th with $(i+1)$th], whereas they ligate every odd numbered block with the next even numbered block [i.e., $(2i-1)$th with $(2i)$th]. In step 3 we hierarchically ligate overlapping adjacent blocks from the previous iteration, until the full sequence is covered (Figure 2, step 3). The ligation strategy is again different from that of Niu et al.'s due to the haplotype consistency constraints imposed by overlapping SNPs, which helps to reduce the candidate haplotype space of the merged blocks. More details about the entire partition-ligation process can be found in the Appendix.

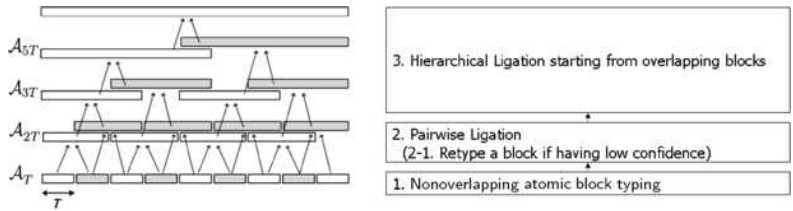

Fig. 2. *The partition-ligation scheme used in Haploi.*

As we reduce the search space based on feasible individual haplotype pairs, there may be possibility of missing some haplotypes in the haplotype space construction if the ligation is only based on disjoint blocks. However, our ligation process considers two blocks with an overlapping region and takes into account all the possible inconsistencies for every heterozygous locus. Therefore, the actual number of haplotypes added to the space can be greater than four in general, except for the first pairwise ligation stage in step 2 (see the Appendix for a detailed example of this). Moreover, even in the pairwise ligation from the nonoverlapping atomic blocks, this risk can be reduced by considering every neighboring pair, not every odd-numbered and even-numbered pair as noted above, as the information in one block can be propagated into both side of neighbors and can be preserved better. Empirically, this new scheme led to a more accurate result than the original PL scheme with greatly improved computational cost, as the original PL scheme cannot be applied to more than a few hundred of the SNPs.

The underlying intuition of our ligation procedure is to allow recombination-like transition on the overlapping regions for including not only all the necessary new haplotype configurations, but also to maximally preserve the haplotypes obtained at previous steps. This heuristic typically results in a population haplotype space of the merged block that is much smaller than the naive product-space of nonoverlapping lower-level blocks. Moreover, individuals whose atomistic haplotypes of the pre-merged blocks have no discrepancy in the overlapping region would not only contribute very few but high-confidence population haplotypes to the pool, but also they need not to be phased again in that ligation step. This constitutes the main source of efficiency and effectiveness of our algorithm.

In summary, comparing to Niu et al.'s PL scheme, our method attempts to build a more parsimonious set of population haplotypes at each ligation iteration by using an individual-based population-haplotype imputation scheme that leverages haplotypic diversity constraints imposed by individual genotypes and overlapping blocks. However, these modifications only help to better trim the population haplotype space; statistically, it results in a near irreducible (due to restriction on the search space) but faster mixing Markov chain during haplotype sampling.

**4. Results.** We evaluated the proposed HDPM model on both simulated genotype data and real genotype sequences from the International HapMap database. The haplotype inference accuracy under HDPM (via the *Haploi* program) is compared to that of the the baseline DP mixture model, and to PHASE 2.1.1 [Stephens, Smith and Donnelly (2001), Stephens and Scheet (2005)], fastPHASE [Scheet and Stephens (2006)], MACH1.0 [Li and Abecasis (2006)] and Beagle 2.1.3 [Browning and Browning (2007)], in their default parameter settings unless otherwise specified. Two different error measures are used: $err_s$, the ratio of incorrectly phased SNP sites over all nontrivial heterozygous SNPs, and $d_w$, the switch distance, which is the number of phase flips required to correct the predicted haplotypes over all nontrivial cases. For short SNP sequences we primarily use $err_s$, whereas for long sequences we compare $d_w$ according to common practice. In addition to haplotype inference, on the simulated data we also estimated other metrics of interest, such as the haplotype frequencies, the mutation rates $\theta$ of each founding haplotypes and the number of reconstructed haplotype founders $K$, to assess the consistency of our model.

4.1. *Simulated multi-population SNP data.* To simulate multi-population genotypes, we used a pool of haplotypes taken from the coalescent-based synthetic dataset in Stephens, Smith and Donnelly (2001), each containing 10 SNPs, as the hypothetical founders; and we drew each individual's haplotypes and genotype by randomly choosing two ancestors from these founders and applying the mutation and noisy genotyping models described in the methodology section. For each of our synthetic multi-population data sets, we simulated five populations, each with 20 individuals. Each population is derived from 5 founders, where two of them are shared across all the populations, and the other three are population-specific. Thus, the total number of founders across the five populations is 17. We test our algorithm on two data sets with different degrees of sequence diversity. In the *conserved* data set we set the mutation rate $\theta$ to be 0.01 for all populations and all loci in the simulation; in the *diverse* data set, $\theta$ is set to be 0.05. All populations and loci are assumed to have the same genotyping error rate. Fifty random samples were drawn from both the conserved and the diverse data sets.

4.1.1. *Haplotype accuracy.* We compare *Haploi* (i.e., HDP) and other methods applied in two modes on synthetic data. Given multi-population genotype data, to use DP or other extant methods, one can either adopt mode-I, pool all populations together and jointly solve a single haplotype inference problem that ignores the population label of each individual, or follow mode-II, apply the algorithm to each population and solve multiple haplotype inference problems separately. *Haploi* takes a different approach, by making explicit use of the population labels and jointly solving multiple coupled haplotype inference problems. Note that when only a single population is concerned, or no population label is available, *Haploi* is still applicable and is equivalent to a baseline DP with one more layer of DP hyper-prior over the base measure. We compare the overall performance of *Haploi* on the whole data with other algorithms run in mode-I, and also the accuracy of *Haploi* within each population with those of other methods run in mode-II. Since fastPHASE can also take account of population labels, we supplied the labels to fastPHASE in mode-I experiments.

We first test how much HDP can gain by the hierarchical structure on multiple populations compared to the baseline DP. Figure 3 compares the result of HDP with the baseline-DP in mode-I (denoted by DP-I) and that in mode-II (denoted by DP-II) on synthetic multiple populations. On both the conserved samples, which are presumably easier to phase, and the diverse samples, which are more challenging, HDP significantly outperformed

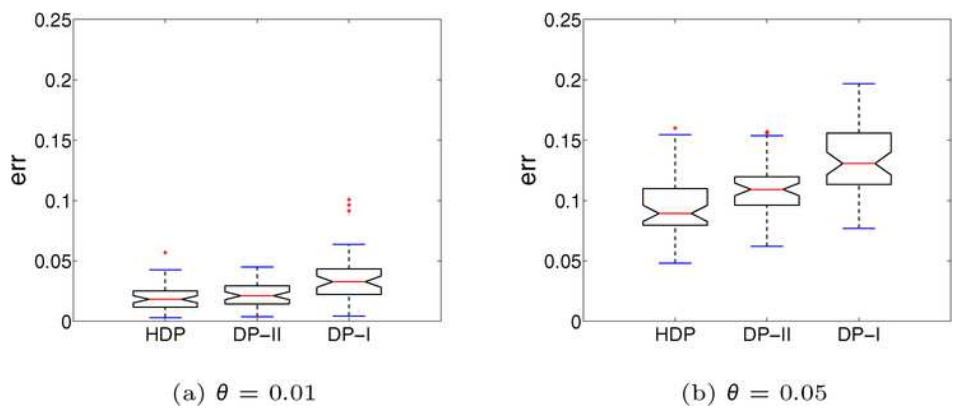

FIG. 3. *A comparison of HDP with the baseline DP on the synthetic multi-population data. DP-II: DP run on each separate population (mode-II). DP-I: DP run on a merged population (mode-I). The errors measured by site-discrepancies over 50 random samples are presented for* (a) *conserved datasets ($\theta = 0.01$) and* (b) *diverse datasets ($\theta = 0.05$).*

DP in both modes (with $p = 0.0336$ against DP-II on the conserved samples, and $p \leq 1.83 \times 10^{-6}$ in all other comparisons, according to a paired $t$-test). In addition, as a baseline case, we applied HDP to each single-population separately as DP in mode-II, assuming the scenario of a single population or individuals without population labels. Again, HDP applied to all populations jointly outperformed this *baseline HDP* significantly, as the latter is deprived of the gain by information sharing. Moreover, this baseline HDP also dominates DP in mode-II significantly, especially on diverse datasets ($p \leq 0.0017$). It appears that the hierarchical structure of HDP which introduces a nonparametric hyper-prior over the base measure of a DPM allows more flexibility in the model and gives better performance than a plain DPM with fixed base measure.

Figure 4 shows boxplots for the differences between the error rate of each benchmark algorithm and that of HDP (i.e., $err_{\{alg\}} - err_{HDP}$). Note that the regions above the horizontal line $y = 0$ correspond to the cases where HDP outperforms others. When other algorithms are run in mode-I [Figure 4(a)], *Haploi* outperforms all of them significantly on both the conserved and diverse samples ($p \leq 8.9 \times 10^{-5}$). *Haploi* remains competitive in comparison with other methods when the latter are run in mode-II, that is, on each population separately [Figure 4(b)]. On the conserved data, PHASE shows the best result, but the differences between algorithms are not significant ($p \leq 0.11$). Whereas on the diverse data, *Haploi* outperforms other algorithms significantly ($p \leq 0.0043$). Again, all significant scores were computed according to a paired $t$-test.

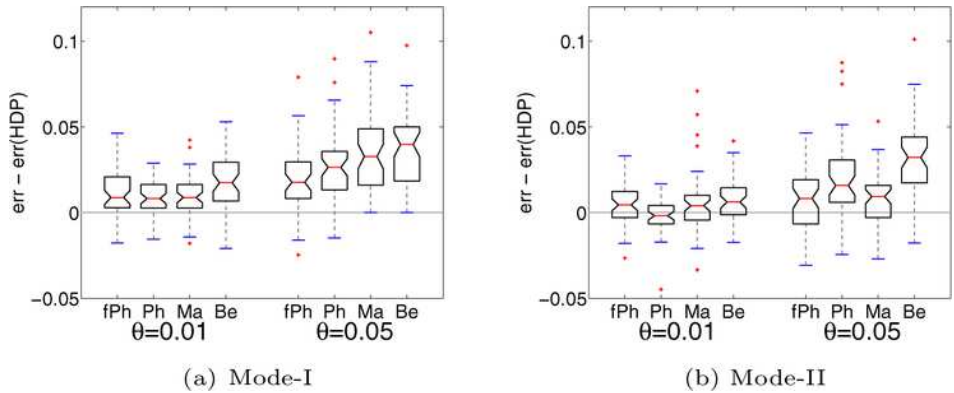

FIG. 4. *A comparison of HDP with other methods (fPh: fastPHASE, Ph: Phase, Ma: Mach, Be: Beagle) running in* (a) *mode-*I*, and* (b) *mode-*II*, on synthetic multi-population data. Boxplots for the differences between the error rate of each algorithm and that of HDP (i.e., $err_{\{alg\}} - err_{HDP}$) are presented.*

4.1.2. *Parameter estimation and sensitivity analysis.* Typically, with random initialization, the Gibbs sampler for *Haploi* converges within 1000 iterations on the synthetic data. This contrasts sampling algorithms used in some of the other haplotype models, which typically need tens of thousands of iterations to reach convergence. The fast convergence is possibly due to *Haploi*'s ability to quickly infer the correct number of founding haplotypes underlying the genotypes samples, which leads to a model significantly more compact (i.e., parsimonious) than that derived from other methods.

*Estimating $K$ and $\theta$.* We compared the estimated $K$—the number of recovered ancestors via both HDP and DP. Recall that we expect $K$ to be 17. Overall, the estimated $K$ under both the DP and HDP models turns out to be very close to this number on the *conserved* datasets; from the diverse data sets, HDP can still offer a good estimate of the number of ancestors, whereas DP recovered more ancestors (around 25 on average) than the true number. This is not surprising since a haplotype which appears in more than one population can have different frequencies in different populations, the baseline DP cannot capture such sub-population structure, and the higher divergence due to both mutation and population diversification can make it generate more ancestors to describe the given dataset.

Our Gibbs sampler also provides reasonable estimates of the mutation rates of each haplotype founder. We observe that for the conserved data sets, HDP yields highly consistent and low variance estimations of $\theta$, and the quality of the estimates due to DP is slightly worse. For the diverse data both algorithms tend to slightly underestimate the mutation rates, and the variance is also higher. It is noteworthy that, in principal, high haplotype diversity of a population can be explained by two competing sources: high mutation rate from ancestors to descendants, and large number of ancestors. Indeed, $K$ and $\theta$ cannot be independently determined, following a similar argument of the un-identifiability of the evolution time and population size under the IAM model. But empirically, HDP appears to strike a reasonable balance between $K$ and $\theta$, and offers plausible estimates of both.

A more thorough sensitivity analysis with respect to the hyper-parameters in our model is detailed in Table 1. The proposed HDP model has two scale parameters, $\gamma$ and $\tau$, for the upper and lower level DP, which are under inverse Gamma priors as discussed in Section 2.2.3. To see the sensitivity of the $K$ and $\theta$ estimations under different priors, we applied various values of hyper parameters $\iota$ and $\kappa$ (the same for both $\gamma$ and $\tau$) on one of the 50 random conserved datasets. Columns 4–9 in Table 1 show the number of recovered founders within each sub-population (the correct number is 5 for each), and the total number of distinct founders over all the populations. Overall, over a wide range of values for the hyper-parameters, *Haploi* gives low-bias and low-variance estimation of the number of founders of each sub-population as well as the total number of distinct founders. In columns 10–11, we show the inferred mutation rate and the haplotyping error. Even when incorrect numbers of founders are recovered, the actual haplotyping errors are not significantly affected, which shows the robustness of the proposed approach for ahaplotype recovering application. The test on a diverse dataset shows a similar tendency while the result is slightly less stable (see supplementary material [Sohn and Xing (2008)] for more details).

TABLE 1
*A sensitivity analysis to the hyper-parameters of HDP on a conserved dataset. Results with different hyper-parameters $\iota$ and $\kappa$ for inverse Gamma prior are shown. The number of founders for each population $(K_i)$ and the total number of ancestors across all the populations are shown in columns 4–9. The estimated mutation rate $\theta$ and the haplotyping errors $(err_s)$ are also shown through columns 10–11. The sensitivity of $\theta$ estimate to the hyper prior is examined over a wide range of both different magnitudes (0.1 to 1000) and ratios (0.0001 to 10,000) of $\iota$ and $\kappa$*

| $\kappa$ | $\iota$ | $\kappa/\iota$ | $K_1$ | $K_2$ | $K_3$ | $K_4$ | $K_5$ | total $K$ (17) | $\theta$ (0.005) | $err_s$ |
|---|---|---|---|---|---|---|---|---|---|---|
| 0.1 | 0.1 | 1 | 5.0 | 5.0 | 5.0 | 5.0 | 5.0 | 17.8 | 0.005 | 0.0058 |
| | 0.5 | 0.2 | 5.0 | 5.0 | 5.0 | 5.0 | 5.0 | 17.5 | 0.004 | 0.0116 |
| | 1 | 0.1 | 5.0 | 5.0 | 5.0 | 5.0 | 5.0 | 18.0 | 0.004 | 0.0000 |
| | 10 | 0.01 | 5.0 | 5.0 | 5.0 | 5.0 | 5.0 | 18.0 | 0.004 | 0.0087 |
| | 100 | 0.001 | 5.0 | 4.0 | 5.0 | 5.0 | 4.0 | 16.0 | 0.007 | 0.0029 |
| | 1000 | 0.0001 | 5.0 | 5.0 | 5.0 | 5.0 | 4.0 | 17.0 | 0.004 | 0.0029 |
| 0.5 | 0.1 | 5 | 5.0 | 5.1 | 5.0 | 5.0 | 5.0 | 18.1 | 0.004 | 0.0087 |
| | 0.5 | 1 | 5.0 | 4.1 | 5.0 | 5.0 | 5.0 | 17.1 | 0.007 | 0.0029 |
| | 1 | 0.5 | 5.0 | 5.0 | 5.0 | 5.0 | 5.0 | 18.0 | 0.004 | 0.0029 |
| | 10 | 0.05 | 5.0 | 5.0 | 5.0 | 5.0 | 5.0 | 18.0 | 0.004 | 0.0145 |
| | 100 | 0.005 | 5.0 | 5.0 | 5.0 | 5.0 | 4.0 | 17.0 | 0.004 | 0.0029 |
| | 1000 | 0.0005 | 5.0 | 5.0 | 5.0 | 5.0 | 4.0 | 17.0 | 0.005 | 0.0087 |
| 1 | 0.1 | 10 | 5.0 | 5.0 | 5.0 | 6.0 | 5.0 | 18.0 | 0.006 | 0.0116 |
| | 0.5 | 2 | 5.0 | 5.0 | 5.0 | 5.0 | 5.0 | 18.0 | 0.004 | 0.0058 |
| | 1 | 1 | 5.0 | 5.0 | 5.0 | 5.0 | 5.0 | 18.0 | 0.004 | 0.0087 |
| | 10 | 0.1 | 5.0 | 5.0 | 5.0 | 5.0 | 5.0 | 18.0 | 0.004 | 0.0029 |
| | 100 | 0.01 | 5.0 | 4.0 | 5.0 | 5.0 | 4.0 | 16.0 | 0.007 | 0.0087 |
| | 1000 | 0.001 | 5.0 | 4.9 | 5.0 | 5.0 | 4.0 | 16.9 | 0.005 | 0.0087 |
| 10 | 0.1 | 100 | 5.0 | 5.0 | 5.0 | 5.3 | 5.0 | 17.1 | 0.004 | 0.0000 |
| | 0.5 | 20 | 5.0 | 5.0 | 5.0 | 5.0 | 5.0 | 18.0 | 0.004 | 0.0087 |
| | 1 | 10 | 5.0 | 5.0 | 5.0 | 5.0 | 5.0 | 18.1 | 0.004 | 0.0029 |
| | 10 | 1 | 5.0 | 5.0 | 5.0 | 5.0 | 5.0 | 18.0 | 0.004 | 0.0000 |
| | 100 | 0.1 | 5.0 | 4.0 | 5.0 | 5.0 | 5.0 | 17.0 | 0.007 | 0.0058 |
| | 1000 | 0.01 | 5.0 | 5.0 | 5.0 | 5.0 | 4.0 | 17.0 | 0.004 | 0.0087 |
| 100 | 0.1 | 1000 | 5.8 | 5.5 | 5.6 | 6.1 | 6.0 | 18.2 | 0.010 | 0.0116 |
| | 0.5 | 200 | 5.2 | 5.2 | 5.2 | 5.8 | 5.5 | 18.4 | 0.008 | 0.0116 |
| | 1 | 100 | 5.1 | 6.2 | 5.4 | 5.5 | 5.2 | 17.3 | 0.006 | 0.0087 |
| | 10 | 10 | 5.0 | 5.0 | 5.1 | 5.0 | 5.1 | 18.1 | 0.005 | 0.0029 |
| | 100 | 1 | 5.0 | 5.0 | 5.0 | 5.0 | 5.0 | 18.0 | 0.004 | 0.0000 |
| | 1000 | 0.1 | 5.0 | 5.0 | 5.0 | 5.0 | 4.0 | 17.0 | 0.004 | 0.0000 |
| 1000 | 0.1 | 10,000 | 6.8 | 6.3 | 8.5 | 6.0 | 10.3 | 25.6 | 0.003 | 0.0087 |
| | 0.5 | 2000 | 7.1 | 7.0 | 7.4 | 6.6 | 8.5 | 24.5 | 0.006 | 0.0116 |
| | 1 | 1000 | 6.4 | 6.5 | 7.7 | 6.4 | 8.4 | 22.8 | 0.005 | 0.0145 |
| | 10 | 100 | 5.3 | 6.5 | 6.3 | 5.8 | 7.0 | 17.8 | 0.010 | 0.0260 |
| | 100 | 10 | 5.1 | 5.1 | 5.0 | 5.0 | 5.1 | 18.1 | 0.005 | 0.0087 |
| | 1000 | 1 | 5.0 | 5.0 | 5.0 | 5.0 | 5.0 | 18.0 | 0.004 | 0.0029 |

*Estimating haplotype frequencies.* Figure 5 summarizes the accuracy of population haplotype frequencies estimated by each algorithm. The discrepancy between the true frequencies and estimated ones is measured by the KL-Divergence $D_{\mathrm{KL}}(p||q) = \sum_x p(x) \log \frac{p(x)}{q(x)}$. The top row shows the accuracy of HDP along with those of DP in mode-II and in mode-I, and the bottom row shows the differences between the error rates of benchmark algorithms and those from HDP. The left column of Figure 5(a) reports $D_{\mathrm{KL}}$ computed on ALL haplotypes frequencies estimated by different algorithms from the conserved data sets and the right column of Figure 5(a) shows the result when measured only on the frequent haplotypes (i.e., with frequencies $\geq 0.05$). Comparing to the baseline-DP, HDP is as accurate when only frequent haplotypes are considered. When all the frequencies are considered, however, the margin of HDP over DP becomes significant, especially on the diverse dataset ($p = 0.0009$). Overall, *Haploi*, PHASE and MACH work equally well without significant difference in performance on conserved datasets. For more difficult diverse data sets [Figure 5(b)], HDP achieves the lowest discrepancy by a significant margin over all the other algorithms. The runner-up, PHASE, beats fastPHASE and MACH with a small margin. When measured only on the frequent haplotypes [i.e., the right column of Figure 5(b)], the discrepancies decrease significantly, but the relative ordering of all the compared algorithms remains similar, except that now fastPHASE outperforms PHASE ($p = 0.0036$).

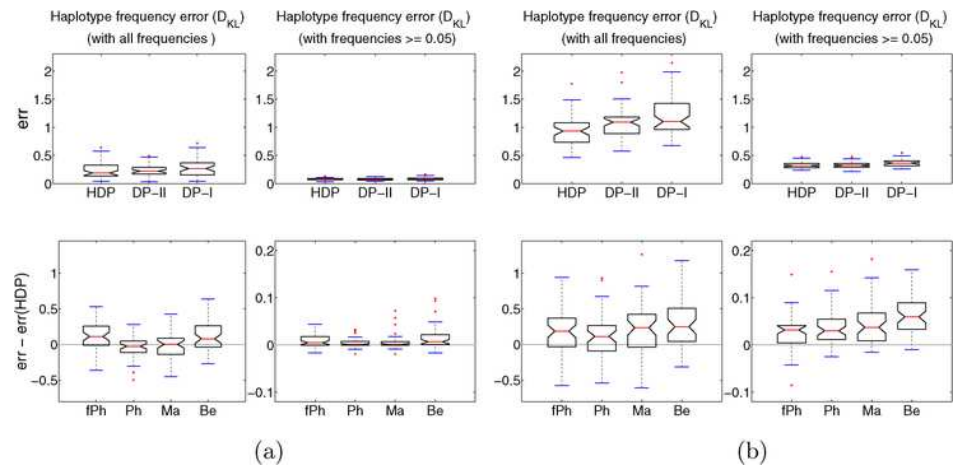

FIG. 5. *A comparison of the accuracies of haplotype frequencies. Top: the result from HDP, DP in mode-II (DP-II), and DP in mode-I (DP-I). Bottom: the relative error rates of four benchmark algorithms with respect to those from HDP.* (a) *Box-plots of $D_{\mathrm{KL}}$'s estimated from the conserved data sets. Left column shows measurements on all haplotypes, right column shows measurements on only the frequent haplotypes.* (b) *Same measurements on the diverse datasets.*

4.2. *The HapMap data.* We also test *Haploi* on both short SNP segments (i.e., ∼6 SNPs) and long SNP sequences (i.e., ∼$10^2$–$10^3$ SNPs) available from the International HapMap Project. This data contains SNP genotypes from four populations: Utah residents with ancestry from northern and western Europe (CEU); Yoruba in Ibadan, Nigeria (YRI); Han Chinese in Beijing (CHB); and Japanese in Tokyo (JPT), with 60, 60, 45 and 44 unrelated individuals, respectively. Although haplotype inference can be, and in some test scenarios, was performed on all populations, evaluation of the outcome is on only the CEPHs and Yorubas since the true haplotypes can be almost unambiguously deduced from trios only in these two populations. The individual genotypes that cannot be unambiguously phased from the trios were ignored in the scoring. We consider three different population-composition scenarios in our experiments below: (1) using all the four populations together for haplotype inference (FourPop); (2) using only CEPH and Yoruba populations for inference (TwoPop); and (3) phasing CEPH and Yoruba separately (OnePop). Essentially, in the FourPop and TwoPop scenarios we solve a bigger haplotype inference problem on data that contain richer population information.

4.2.1. *Short SNP sequences.* Phasing short SNPs is the basic operation of large-scale haplotype inference problems which rely either on partition-ligation heuristics or on model-based methods, such as recombination process, to integrate short phased haplotype segments into long haplotypes. Figure 6 shows a comparison of the phasing accuracy on 6-SNP segments [following a recommendation in Niu et al. (2002) on the optimal size-range of basic units for subsequent ligation] by four algorithms. The test was done on randomly selected 100 sets of 6-SNPs segment from chromosome 21. For each of the three population-composition scenarios, we applied all methods to different population sizes, that is, 60, 30, 20 and 10 individuals per population, to examine the effect of population size on phasing accuracy.

Several aspects of *Haploi*'s performance on real data are revealed by Figure 6. First, comparing the performances of *Haploi* under the three different population-composition scenarios, we observe that *Haploi* improves steadily as more populations are included in haplotype inference, and the improvements are statistically significant. The $p$-values of the differences between FourPop and OnePop scenarios are 0.00024, 0.000038, 0.0016 and 0.000022 for data with 60, 30, 20 and 10 individuals per population, respectively; and the $p$-values of the margins of TwoPop over OnePop are 0.0014, 0.0002, 0.0053 and 0.00047, respectively, in the same order. The improvement in FourPop over TwoPop is less significant, with $p$-values 0.35, 0.11, 0.16 and 0.023, respectively, suggesting that the possible gain in haplotype accuracy enabled by the HDP model via exploring shared information among populations can be capitalized the most when we change from single-population inference to joint-inference in multiple population, whereas the effect of having more populations in the multi-population scenario appears to be less obvious in this dataset.

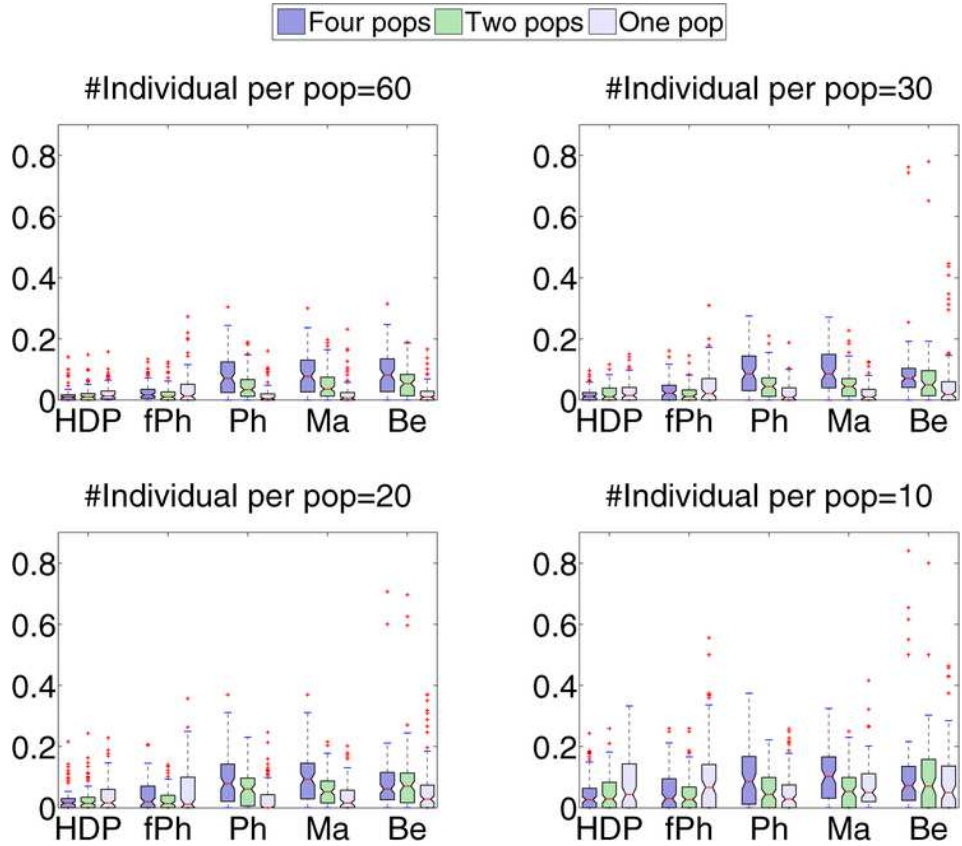

Fig. 6. *A comparison of the haplotyping error on the CEPH + Yoruba population over randomly chosen 100 sets of 6-SNP segments from Chromosome 21. The results were obtained under three population-composition scenarios:* (i) *FourPops: when data from all the four populations were used (blue) for inference;* (ii) *TwoPops: when data from CEPH and Yoruba populations were used together (green);* (iii) *OnePop: when each of the CEPH and Yoruba populations was used separately (gray). Different sample sizes, with 60, 30, 20 and 10 individuals per each population, were used.*

Second, comparing the performances of *Haploi* under different population sizes, we observe that the performance-gain through information sharing among populations tends to be greater when the population sizes decrease. For example, the performance differences of *Haploi* in multi-population over single-population become most significant when the number of individuals per population is the smallest (#Individual per pop = 10). This observation suggests that HDP is especially advantageous under data scarcity situations where information from each population becomes insufficient to warrant reliable inference within the population.

Third, other methods, such as PHASE, MACH and Beagle, appear not able to benefit from increased population diversity as indicated by the significant drop of their accuracies when more populations are involved. The performance of fastPHASE (with known population labels) improves substantially when two populations are used together, while the performance becomes slightly worse in the case of four populations. Comparing the results from the most preferred scenario of each algorithm, that is, *Haploi* under FourPop, fastPHASE under TwoPop, and all the others under OnePop, *Haploi* and PHASE worked similarly well when all the available data were used (i.e., #Individual per pop = 60), with mean error rate of each algorithm at 0.0174, 0.0198, 0.0173, 0.0229 and 0.0222, respectively (with $p = 0.05$, 0.89, 0.10, 0.01 over differences of *Haploi* with other algorithms). When the population sizes decrease, *Haploi* starts to surpass others more substantially, and works more reliably than others. For example, on 10 individuals per population, the mean error rates of the five algorithms were 0.0424, 0.0460, 0.0512, 0.0777 and 0.0945, and the $p$-values of the margin of *Haploi* over others are 0.17, 0.02, $1.2 \times 10^5$, $6.7 \times 10^6$, respectively.

4.2.2. *Long SNP sequences.* Finally we test *Haploi* on very long genotype sequences with $10^2 \sim 10^3$ SNPs. We selected 10 ENCODE regions from the HapMap DB, each spanning roughly 500 Kb and containing from 254 to 972 common SNPs across all four populations (see supplementary material [Sohn and Xing (2008)] for more details). We performed haplotype inference under three different population-composition scenarios as before, but due to the extremely high cost in computational time in these experiments, we only worked on the full-size data sets. Figure 7 shows a comparison of haplotype reconstruction quality, using PHASE, fastPHASE, MACH, Beagle and *Haploi* equipped with the PL heuristic. Out of the 30 experiments we performed (10 regions and three scenarios), the PHASE program failed to yield results in 5 experiments after a 31-day runtime, so we omit the corresponding results in our summary figure.

The conclusion from Figure 7 is less clear than the ones from previous sections from experiments on short SNP sequences and on simulation data. Overall, Beagle dominates all the algorithms with a small margin, PHASE also shows comparable result to Beagle when converged, but all the other algorithms work comparably in most cases across different datasets and different scenarios. In terms of computational cost, Beagle was the fastest, it took less than a minute for each task; fastPHASE and MACH mostly took less than 1 hour for each task, *Haploi* took from 1–10 hours, depending on the length of the sequence, whereas PHASE took one to two orders of magnitude longer, and was indeed impractical for phasing very long sequence.

In summary, our result shows that *Haploi* is competent and robust for phasing long SNP sequences from diverse genetic origins at reasonable time cost, even though it has not yet employed any sophisticated way for processing long sequences, such as the recombination process. Since *Haploi* appeared to outperform other methods over short SNPs, we believe that the competence of *Haploi* on long SNPs is due to a better inference power endowed by the HDP model for multi-population haplotypes, and we expect that an upgrade that incorporates explicit recombination models in conjunction with HDP for long SNPs is likely to lead to more accurate haplotype reconstructions.

**5. Discussion.** We have proposed a new Bayesian approach to haplotype inference for multiple populations using a hierarchical Dirichlet process mixture. By incorporating an HDP prior which couples multiple heterogeneous populations and facilitates sharing of mixture components (i.e., haplotype founders) across multiple Dirichlet process mixtures, the proposed method can infer the true haplotypes in a multi-subpopulation dataset with an accuracy superior to the state-of-the-art haplotype inference algorithms.

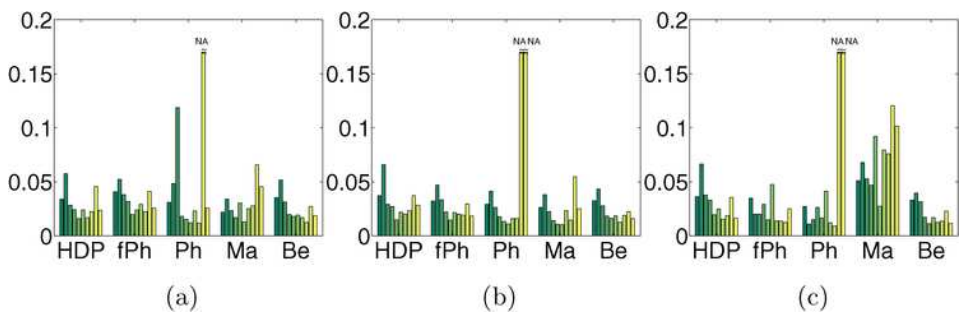

Fig. 7. *Performance on the full sequences of the selected ten ENCODE regions.* (a) *Error rates under four-population scenario.* (b) *Under the two-population scenario.* (c) *Under the one-population scenario. For cases of which the program does not converge (NC) within a tolerable duration (i.e., 800 hours), we cap the bar with a "≉" to indicate that the results are not available (NA).*

Recently, there emerged new models related to our HDP model, the closest being the nested Dirichlet process (NDP) by Rodriguez, Dunson and Gelfand (2006). In an NDP, instead of using a hyper-DP as a common base measure as in HDP to allow sharing of founders across populations, the population-specific DPs are directly drawn from a prior DP, so that not only the founders, but also their frequencies can be shared across populations. Although this model can be more expressive in many applications, it may be less appropriate than HDP for multi-population haplotype problems where excessive structural sharing across populations are not warranted, especially when different populations bear very distinct demography and genetic prototypes. Another strategy proposed by Muller, Quintana and Rosner (2004) employs an explicit stochastic convex combination of a population-specific prior and a universal prior for each founder. Under such a model, once a founder is destined to be shared across populations, it will appear with equal frequency in all populations. HDP subsumes this scenario, but also allows more flexible sharing of the founders.

The proposed model achieves the desirable properties of PAC regarding mutation dynamics [Li and Stephens (2003)], including the parental-dependent-mutation effect, albeit in a very different way. For example, to see the PDM property, note that when the next haplotype is to be sampled according to equation (1), we pick an ancestor of some previously drawn haplotypes, and apply a mutation process to the ANCESTOR (rather than to one of the previously drawn haplotypes as in PAC). This operation implicitly results in a PDM effect among haplotypes by relating them to their corresponding founder via a tractable star genealogy equipped with a common mutation process $P_h(\cdot|founder)$. A new haplotype generated from this process will bear mutations over its corresponding founders rather than being completely random. Above these founders, we model their genealogy and type history by a *coalescent-with-IMA* model, whose resulting marginal is equivalent to that of the Dirichlet process. Here a new founder can be sampled independently of the type-history in the coalescent from the base measure, rather than according to a PDM, with probability proportional to the IMA mutation rate. Putting everything together, the DP mixture model essentially implements a combination of IMA and PDM: it models the genealogy and type history of hypothetical ancestors presumably corresponding to a bottleneck with a coalescent-with-IMA model; below the bottleneck, it uses multiple (indeed, can be countably infinite many) star genealogies rooted at the ancestors present in the bottleneck and equipped with ancestor-dependent Poisson mutation process, to approximate the coalescent-with-PDM model. The time of the bottleneck depends on the value of the scaling parameter $\alpha$ of the DP. One can introduce a prior to this parameter so that it can be estimated a posteriori from data.

It is well known that under Kingman's $n$-coalescent, a dominant portion of the depth of the coalescent tree is spent waiting for the earliest few lineages to coalesce to the MRCA and the majority of lineages of even a very large population can actually coalesce very rapidly into a few ancestors, which means that the net mutation rates from each of these ancestors to their descendants in a modern haplotype sample do not vary dramatically among the descendants. Thus, qualitatively, a star genealogy provides a reasonable approximation to the actual (heavily time-compressed) genealogy of a modern haplotype sample up to these ancestors. As a reward of such approximation, a well-known property of DP mixture is that it defines an exchangeable distribution of the samples. Furthermore, the Pólya urn construction of DP enables simple and efficient Monte Carlo for posterior inference of haplotypes and other parameters of interest, and the DPM formalism offers a convenient path for extensions that capture more complex demographic and genetic scenarios of the sample, such as the multi-population haplotype distribution as we explored in this paper.

Unlike the models underlying PHASE and fastPhase, the PL heuristic used in the *Haploi* program does not explicitly model the recombination process that shapes the LD patterns of long SNP sequences. Since an HDP model without the aid of PL-scheme dominates PHASE and fastPhase over short SNPs, we believe that an upgrade that incorporates an explicit recombination model in conjunction with HDP is likely to lead to more accurate reconstruction of long haplotypes. The hidden Markov Dirichlet process recently developed by us to model recombination in open ancestral space offers a promising path for such an upgrade [Xing and Sohn (2007)]. Under the proposed statistical framework for modeling haplotype and genotype distribution, it is also straightforward to handle various missing value problems in a principled way. In another possible extension, although in the present study we have assumed that the subpopulations' labels of individuals are known, it is straightforward to generalize our method to situations in which the subpopulations' labels are unknown and to be inferred. This opens the door to applications of our method to large-scale genetic studies involving joint inference over markers and demography. The HDP model is also a natural formalism for applications outside of population genetics, such as in text modeling, where one can use an HDPM to model co-clustering of documents from different journals (analogous to different populations here) according to both shared and unique topics defined by, for example, a latent Dirichlet allocation model [Blei, Ng and Jordan (2003)], and also in network modeling, where the neighbor profiles of every node can be modeled by a low-level DPM whose likelihood function is defined by, for example, a mixed membership stochastic block model [Airoldi et al. (2006)], and the entire network corresponds to an HDP over all nodes. Due to space limits, a detailed description of our work in these applications is beyond the scope of this paper.

## APPENDIX A: MARKOV CHAIN MONTE CARLO FOR HDP

In this section we describe a Gibbs sampling algorithm for posterior inference of haplotypes under the HDPM model. We start with a brief description of the HDP formalism in terms of Pólya urn models. Imagine we set up a single "stock" urn at the top level, which contains balls of colors that are represented by at least one ball in one or multiple urns at the bottom level. At the bottom level, we have a set of *distinct* urns which are used to define the DP mixture for each population. Now let us suppose that upon drawing the $m_j$th ball for urn $j$ at the bottom, the stock urn contains $n$ balls of $K$ distinct colors indexed by an integer set $\mathcal{C} = \{1, 2, \ldots, K\}$. Now we either draw a ball randomly from urn $j$, and place back two balls both of that color, or with some probability we return to the top level. From the stock urn, we can either draw a ball randomly and put back two balls of that color in the stock urn and one in $j$, or obtain a ball of a new color $K+1$ with probability $\frac{\gamma}{n-1+\gamma}$ and put back a ball of this color in both the stock urn and urn $j$ of the lower level. Essentially, we have a master DP (the top urn) that serves as a source of atoms for $J$ child DPs (bottom urns).

Associating each color $k$ with a random variable $\phi_k$ whose values are drawn from the base measure $F$, we know that draws from the stock urn can be viewed as marginals from a random measure distributed as a Dirichlet Process $Q_0$ with parameter $(\gamma, F)$. From equation (1), for $n$ random draws $\boldsymbol{\phi} = \{\phi_1, \ldots, \phi_n\}$ from $Q_0$, the conditional prior for $(\phi_n | \boldsymbol{\phi}_{-n})$, where the subscript "$-n$" denotes the index set of all but the $n$th ball, is

$$\phi_n | \boldsymbol{\phi}_{-n} \sim \sum_{k=1}^{K} \frac{n_k}{n-1+\gamma} \delta_{\phi_k^*}(\phi_n) + \frac{\gamma}{n-1+\gamma} F(\phi_i),$$

where $\phi_k^*, k=1, \ldots, K$, denotes the $K$ distinct values (i.e., colors) of $\boldsymbol{\phi}$ (i.e., all the balls in the stock urn), and $n_k$ denotes the number of balls of color $k$ in the top urn.

Conditioning on $Q_0$ (i.e., using $Q_0$ as an atomic base measure of each of the DPs corresponding to the bottom-level urns), the $m_j$th draws from the $j$th bottom-level urn are also distributed as marginals under a Dirichlet measure which leads to the distribution shown in equation (2).

This nested Pólya urn scheme motivates an efficient and easy-to-implement MCMC algorithm to sample from the posterior associated with HDPM. Recall that the mixture components $\phi_k$ correspond to the ancestral haplotypes with their mutation rates, and the samples correspond to individual haplotypes. Therefore, the variables of interest are $a_{k,t}$, $h_{ie,t}^{(j)}$, $c_{ie,t}^{(j)}$, $\gamma$ and $\tau$, and $g_{i,t}^{(j)}$ (the only observed variables). We may assume that the represented mixture components are indexed by $1, \ldots, K$, the weights of the founders at the top level DP is $\beta = (\frac{n_1}{n-1+\gamma}, \ldots, \frac{n_K}{n-1+\gamma}, \frac{\gamma}{n-1+\gamma})$, where $\frac{\gamma}{n-1+\gamma}$ is the total weight corresponding to some unrepresented founder $K+1$, and the weights of founders at the bottom-level DP for, say, the $j$th population, are $(\frac{m_{j,1}}{m_j-1+\tau},\ldots,\frac{m_{j,K}}{m_j-1+\tau},\frac{\tau}{m_j-1+\tau})$, where $\frac{\tau}{m_j-1+\tau}$ corresponds to the probability of consulting the top-level DP. The Gibbs sampler alternates between three coupled stages. First, we sample the scaling parameters $\gamma$ and $\tau$ of the DPs according to equation (7).

Then, we sample the $c_{i_e}^{(j)}$ and $a_{k,t}$ given the current values of the hidden haplotypes and the scaling parameters. Before sampling $c_{i_e}^{(j)}$, we first erase its contribution to the sufficient statistics of the model. If the old $c_{i_e}^{(j)}$ was $k'$, set $m_{jk'} = m_{jk'} - 1$. If it was sampled from the top level DP, we also set $n_{k'} = n_{k'} - 1$. Note that $c_{i_e}^{(j)} \leq K+1$ (i.e., indicating existing founders, plus a new one to be instantiated). Now we can sample $c_{i_e}^{(j)}$ from the following conditional distribution:

$$
\begin{aligned}
p(c_{i_e}^{(j)} &= k | \mathbf{c}^{[-j,i_e]}, \mathbf{h}, \mathbf{a}) \\
&\propto p(c_{i_e}^{(j)} = k | \mathbf{c}^{[-j,i_e]}, \mathbf{m}, \mathbf{n}) p(h_{i_e}^{(j)} | a_k, \mathbf{c}, \mathbf{h}^{[-j,i_e]}) \\
&\propto (m_{jk}^{[-j,i_e]} + \tau \beta_k) p(h_{i_e}^{(j)} | a_k, \mathbf{l}_k^{[-j,i_e]}) \qquad \text{for } k=1,\ldots,K+1,
\end{aligned}
\tag{8}
$$

where $m_{jk}^{[-j,i_e]}$ represents the number of $c_{i'_{e'}}^{(j)}$ that are equal to $k$, except $c_{i_e}^{(j)}$ in group $j$, and $m_{j,K+1} = 0$; $\mathbf{l}_k^{[-j,i_e]}$ denotes the sufficient statistics associated with all haplotype instances originating from ancestor $k$, except $h_{i_e}^{(j)}$. If, as a result of sampling $c_{i_e}^{(j)}$, a formerly represented founder is left with no haplotype associated with it, we remove it from the represented list of founders. If, on the other hand, the selected value $k$ is not equal to any other existing index $c_{i_e}^{(j)}$, that is, $c_{i_e}^{(j)} = K+1$, we increment $K$ by 1, set $n_{K+1} = 1$, update $\beta$ accordingly, and sample $a_{K+1}$ from its base measure $F$.

Now, from equation (4), we can use the following posterior distribution to sample $a_k$:

$$
\begin{aligned}
p(a_{k,t} | \mathbf{c}, \mathbf{h}) &\propto \prod_{j, i_e | c_{i_e,t}^{(j)} = k} p(h_{i_e,t}^{(j)} | a_{k,t}, l_{k,t}^{(j)}) \\
&= \frac{\Gamma(\alpha_h + l_{k,t}) \Gamma(\beta_h + l'_{k,t})}{\Gamma(\alpha_h + \beta_h + m_k)(|\mathcal{A}|-1)^{l'_{k,t}}} R(\alpha_h, \beta_h),
\end{aligned}
\tag{9}
$$

where $l_{k,t}$ is the number of allelic instances originating from ancestor $k$ at locus $t$ across the groups that are identical to the ancestor, when the ancestor has the pattern $a_{k,t}$. If $k$ was not represented previously, we can just use zero values of $l_{k,t}$, which is equivalent to using the probability $p(a|h_{i_e}^{(j)})$.

We now proceed to the third sampling stage, in which we sample the haplotypes $h_{i_e}^{(j)}$, given the current state of the ancestral pool and the ancestral haplotype assignment for each individual, according to the following conditional distribution:

$$
\begin{aligned}
p(h_{i_e,t}^{(j)}|\mathbf{h}_{[-i_e,t]}^{(j)},\mathbf{c},\mathbf{a},\mathbf{g}) &\propto p(g_{i,t}^{(j)}|h_{i_e,t}^{(j)},h_{i_{\overline{e}},t}^{(j)},\mathbf{u}_{[-i_e,t]}^{(j)})p(h_{i_e,t}^{(j)}|a_{k',t},\mathbf{l}_{k',[-i_e,t]}^{(j)}) \\
&= R_g \frac{\Gamma(\alpha_g + u)\Gamma(\beta_g + (u' + u''))}{\Gamma(\alpha_g + \beta_g + IJ)}[\mu_1]^{u'}[\mu_2]^{u''} \\
&\quad \times R_h \frac{\Gamma(\alpha_h + l_{k',i_e,t}^{(j)})\Gamma(\beta_h + l_{k',i_e,t}^{\prime(j)})}{\Gamma(\alpha_h + \beta_h + n_k)(|\mathcal{A}| - 1)^{l_{k',i_e,t}^{\prime(j)}}},
\end{aligned}
$$
(10)

where $k' \equiv c_{i_e}^{(j)}$, $l_{k,i_e,t}^{(j)} = l_{[-i_e,t]}^{(j)} + \mathbb{I}(h_{i_e,t}^{(j)} = a_{k,t})$, and $\mathbf{u}_{[-i_e,t]}^{(j)}$ are the set of sufficient statistics recording the inconsistencies between the haplotypes and genotypes in population $j$.

## APPENDIX B: DETAILS OF THE PL PROCEDURE

This section describes the detailed procedure of partition-ligation algorithm used in *Haploi*, which can be divided into three steps: (1) atomic block typing; (2) bottom-level pairwise ligation to generate overlapping blocks; and (3) hierarchical ligation of overlapping blocks until only one block is left. In step 1, we partition given genotype sequences into $L$ short blocks of length $T$ and phase each atomistic block using the proposed HDPM. From this step, we obtain all the individual haplotypes and also the population haplotype pool for each block. Let $\mathcal{A}_i^T \equiv \{A_{k,(i-1)T+1:\,iT}|k=1,\ldots,K_i^T\}$ denote the population haplotype pool for $T$ SNPs in the $i$th block which ranges from locus $(i-1)T+1$ to $iT$.

In the next step, we ligate every pair of neighboring blocks: $\mathcal{A}_i^T \& \mathcal{A}_{i+1}^T \to \mathcal{A}_i^{2T}, i=1,\ldots,L-1$. Specifically, for each pair of neighboring blocks $i$ and $i+1$, given $\mathcal{A}_i^T$ and $\mathcal{A}_{i+1}^T$, we can impute at most four new stitched haplotypes from an individual since each individual has only two possible haplotypes within each block. In practice, we often have fewer because an individual can be homozygous or the stitched haplotype may already have been imputed from earlier individuals. We pool such stitched haplotypes from all the individuals to form $\mathcal{A}_i^{2T}$, which usually leads to only a small subset of $\mathcal{A}_i^T \times \mathcal{A}_{i+1}^T$. Then based on a finite dimensional Dirichlet prior over $\mathcal{A}_i^{2T}$, we do Gibbs sampling as in Niu et al.'s PL scheme to obtain individual haplotypes for each overlapping $2T$ region. To compensate possible ill-ligated blocks, we can redo the direct haplotype inference based on HDPM on those merged blocks whose entropy of haplotype distribution is above some threshold (Figure 2, step 2–1). This is computationally affordable since the length of the ligated block at this stage is not yet too big and we can start with better initialization than random assignment. The output from step 2 is $L-1$ sets of length $2T$ population haplotypes, $\{\mathcal{A}_i^{2T} : i = 1, \ldots, L-1\}$, overlapping on $T$ loci for each adjacent pair, and all individual haplotypes in these length $2T$ overlapping segments.

In step 3, we hierarchically ligate overlapping adjacent blocks from the previous iteration, until the full sequence is covered (Figure 2, step 3). Specifically, as in step 2, we build the candidate population haplotype pool by adding every unique stitched-haplotype resulted from ligating the haplotypes of the two shorter blocks in every individual. When the overlapping regions of a pair of atomistic haplotypes in an individual are consistent, ligation to a longer haplotype is trivially a merging of the two overlapping haplotypes, and this avoids generating all combinations of the atomistic haplotypes from each block. Only when the overlapping regions in an individual are inconsistent, we grow the haplotype space of the merged blocks by including all possible ligations consistent with the atomistic haplotypes and the individual genotype. For example, suppose a particular individual's haplotypes were recovered as 000**100**/100**010** at loci 1 to 6 for the first block, and **110**000/**000**100 at loci 4 to 9 for the next block, and three SNPs are overlapping in the two blocks. Then to accommodate the discrepancy on the 4th and 5th SNPs, we have four possible haplotypes, 10, 01, 00, 11, for these two loci; for the remaining parts of the region covered by these two blocks, that is, loci 1–3 and loci 6–9, we have two haplotypes (which are from the atomistic haplotypes determined in the previous iteration) for each of them. So a combination of all these possibilities will add the following sixteen haplotypes to the population haplotype space for the ligated segment:

000**100**000/010**010**100, 000**110**000/100**000**100, 000**010**000/100**100**100,
000**000**000/100**110**100, 000**100**100/100**010**000, 000**110**100/100**000**000,
000**010**100/100**100**000, 000**000**100/100**110**000.

Under the newly formed population haplotype space at each ligation iteration, we again apply a Gibbs sampler as in step 2 to determine the individual haplotypes of all remaining unphased individuals over the ligated block under a fixed-dimensional Dirichlet prior of the haplotype frequencies in this trimmed haplotype space. We continue this process hierarchically until there is only one block left. Since each time we only employ overlapping regions of size $T$, the number of steps needed to complete the ligation of a whole sequence is comparable to Niu et al. 's hierarchical PL scheme.

**Acknowledgments.** We would like to thank anonymous reviewers and the associate editor for insightful and detailed comments on our manuscript, which helped to greatly improve this paper.

## SUPPLEMENTARY MATERIAL

**More results on sensitivity analysis and description of real data** (DOI: 10.1214/08-AOAS225SUPPB; .pdf). We provide the sensitivity analysis result to the hyper-parameters of HDP on diverse dataset ($\theta = 0.05$), and the details of the real data from 10 HapMap ENCODE regions used in Figure 7.

School of Computer Science  
Carnegie Mellon University  
Pittsburgh, Pennsylvania 15213  
USA  
E-mail: ksohn@cs.cmu.edu  
epxing@cs.cmu.edu